\documentclass{article}
\usepackage{natbib}
\usepackage{PRIMEarxiv}
\usepackage{amsmath,amsfonts}
\usepackage{algorithmic}
\usepackage{algorithm}
\usepackage{array}
\usepackage[caption=false,font=normalsize,labelfont=sf,textfont=sf]{subfig}
\usepackage{textcomp}
\usepackage{stfloats}
\usepackage{url}
\usepackage{verbatim}
\usepackage{graphicx}
\setcitestyle{authoryear,open={(},close={)}}
\usepackage{cleveref}
\usepackage{csquotes}

\newcommand{\x}{\mathbf{x}}
\newcommand{\w}{\mathbf{w}}
\newcommand{\R}{\mathbb{R}}
\newcommand{\E}{\mathbb{E}}
\renewcommand{\P}{\mathbb{P}}

\newcommand{\X}{\mathcal{X}}
\newcommand{\C}{\mathcal{C}}
\newcommand{\M}{\mathcal{M}}
\newcommand{\D}{\mathcal{D}}
\newcommand{\I}{\mathbb{I}}

\newcommand{\tmax}{{t^m_{\max}}}
\newcommand{\eqd}{\overset{d}{=}}
\DeclareMathOperator*{\argmax}{arg\,max} 

\begin{document}

\title{Comments on \enquote{Surrogate Modeling for Bayesian Optimization Beyond a Single Gaussian Process}}

\author{%
    Jack Sandberg,
  Morteza Haghir Chehreghani \\
  Department of Computer Science and Engineering\\
  Chalmers University of Technology and University of Gothenburg\\
  Gothenburg, Sweden \\
  \texttt{\{jack.sandberg, morteza.chehreghani\}@chalmers.se}
}

\maketitle

\begin{abstract}
\citet{luSurrogate2023} proposed Ensemble Gaussian Process Thompson Sampling (EGP-TS) for Bayesian Optimization and derived bounds on the Bayesian regret in three settings: a standard sequential setting, an asynchronously parallel setting, and a synchronously parallel setting. In this paper, we show that the proof for the standard sequential setting does not hold due to multiple issues and that a recent revision \citep{luSurrogate2026} does not resolve these issues. Consequently, the proofs for the parallel settings are also invalidated. 
\end{abstract}

\section{Introduction}
Bayesian optimization (BO) provides a flexible framework for optimizing unknown functions where (potentially noisy) observations of the function are costly to obtain \citep{garnettBayesian2023}. In BO, the unknown function being optimized is implicitly or explicitly assumed to be sampled from a Gaussian process with a specific mean and kernel function. The choice of kernel function further imposes assumptions on the properties of the unknown function. E.g. the smoothness parameter of the Matérn kernel \citep{maternSpatial1986} determines how many times the unknown function is continuously differentiable \citep{dacostaSample2025} which can significantly impact the difficulty of the optimization problem.    

Theoretical analysis of Bayesian optimization algorithms commonly assume that the mean and kernel function are known. In practice, parameters of the kernel function are often estimated using maximum likelihood allowing the kernel to adapt to the collected data. However, such adaptive methods lack theoretical performance guarantees. 
\citet{luSurrogate2023} proposed Ensemble GP-TS (EGP-TS), a BO algorithm that adaptively selects probable kernel functions, and provided analytical performance guarantees for EGP-TS. EGP-TS combines Thompson sampling with a GP mixture prior. In each iteration, EGP-TS first samples a kernel according to its posterior probability of generating the data and then selects a location to query the unknown function based on the posterior probability of it being the location of the optima given the sampled kernel.

\citet{luSurrogate2023} analyzed the Bayesian regret of EGP-TS and presented three theorems. Theorem 1 states that EGP-TS has sublinear regret in the standard setting without parallelism whereas Theorem 2 and 3 extend this result to the asynchronously and synchronously parallel setting respectively. The proofs rely on the technique of prior confidence sets introduced by \citet{hongThompson2022}. In \citet{hongThompson2022}, the authors introduced MixTS, a bi-level Thompson sampling algorithm for bandit problems with mixture priors for which EGP-TS can be seen as an instantiation of. Recently, \citet{sandbergAdaptive2026} identified issues with the proof of the regret bound for MixTS in the linear setting \citep{hongThompson2022}.

In this paper, we highlight four issues with the proof of Theorem 1 and compare them with the issues of \citet{hongThompson2022}. The first issue appears to be a typo. However, the three remaining issues are related to the novel part of the proof that bounds $\mathcal{BR}_2(T)$ (term $A_4$ in the proof in the appendix) using the techniques of \citet{hongThompson2022}. The issues of \citet{luSurrogate2023} and \citet{hongThompson2022} arise at mostly the same parts of the proofs but with some differences due to variations in the proof techniques. 
Furthermore, we show that a recent revision of the proof \citep{luSurrogate2026} does not resolve these issues.
Consequently, the proofs of all three theorems are invalid. A corrected analysis of the same algorithm is given in \citet[Theorem 4.4]{sandbergAdaptive2026}, where EGP-TS is called HP-GP-TS.

\section{Setup and notation}
We briefly introduce the setup and necessary notation from \citet{luSurrogate2023}. We consider the optimization problem $\x_* = \argmax_{\x \in \X} f(\x)$ in a setting where $f$ is sequentially queried at selected inputs $\x_t$ with noisy observations $y_t = f(\x_t) + \epsilon_t$, $\epsilon_t \sim N(0, \sigma^2_n)$ for $t = 1,\ldots, T$. The prior model $m_*$ is sampled from a categorical distribution $m_* \sim \mathcal{C}AT(\mathcal{M}, \w_0)$ of size $M = |\M|$ and with prior weights $\w_0$. The function $f$ is sampled from a Gaussian process with a kernel function determined by $m_*$: $f \sim \mathcal{GP}(0, \kappa^{m_*})$. The Bayesian cumulative regret is defined as $\mathcal{BR}(T) = \E\left[ \sum_{t=1}^T f(\x_*) - f(\x_t) \right]$ where the expectation is taken w.r.t. all of the randomness in the problem, including $m_*$, $f$, $(x_i, y_i)_{i=1}^T$ and internal randomness in the algorithm. 

During the optimization process, the EGP-TS algorithm first samples a model from the posterior categorical distribution, $m_t \sim \mathcal{C}AT(\M, \w_t)$, then samples a function from the posterior GP distribution of $m_t$, $\tilde f_t(\x) \sim \mathcal{GP}(\mu_{t-1}^{m_t}(\x), \kappa_{t-1}^{m_t}(\x, \x'))$ where $\mu_{t-1}^{m_t}(\x)$ and $\kappa_{t-1}^{m_t}(\x, \x')$ is the posterior mean and kernel function of model $m_t$, and finally queries the black-box function at the input that maximizes the sampled function $\x_t := \arg\max_{\x \in \X} \tilde f_t(\x)$.

The confidence set $\C_t \subseteq \M$, introduced by \citet{hongThompson2022}, is a tool in the regret analysis that tracks which prior models $m \in \M$ have led to observations $\{ y_t : m_t = m\}$ that are not too surprising assuming that $m_* = m$. It is defined as 
\begin{equation}
    \C_t = \{ m \in \M : G_t^m \leq 2 \sigma_n \sqrt{N_{t-1}^m \log T}\}
\end{equation}
where the number of draws of $m$ is $N_{t-1}^m = \sum_{\tau=1}^{t-1} \I(m_\tau = m)$, and the excess reward of $m$ is $G_t^m = \sum_{\tau=1}^{t-1} \I(m_\tau = m) (L_\tau^m(\x_\tau) - y_\tau)$. The lower confidence bound for model $m$ is $L_t^m(\x) = \mu_{t-1}^m(\x) - \eta \sigma_{t-1}^m(\x)$ with $\eta = 2 \sqrt{\log T}$.

\section{Issues in the proofs of \citet{luSurrogate2023}}
In this section, we present the identified issues in the regret bounds of EGP-TS in the standard BO setting and discuss how they affect the parallel setting. 
\subsection{Lemma 3 statement} \label{sec:lem3}
In Lemma 3, it is stated that $\max_{\mathbf{x} \in \X} |f(\x_*) - f(\x)| \leq 2B $ where $B = \E[\sup_{\x \in \X}|f(x)|]$ for $f \sim \mathcal{GP}(0, \kappa^m)$. 
However, $f(\x_1) - f(\x_2)$ is Gaussian, and in Bayesian optimization it reasonably has positive variance for some pair $(\x_1, \x_2)$ (otherwise $f$ is almost surely constant). Since $\max_{\mathbf{x} \in \X} |f(\x_*) - f(\x)| \geq |f(\x_1) - f(\x_2)|$, the maximum cannot be bounded by a constant w.p. 1, and thus the statement of the lemma is false. 
The lemma would be correct if stated with an expectation around $\max_{\mathbf{x} \in \X} |f(\x_*) - f(\x)|$.

\subsection{Proof of Lemma 5} \label{sec:lem5}
Lemma 5 claims that $\E[\mu_{t-1}^{m_t}(\x_t) - f(\x_t)] \leq 2B$. 
In the proof of Lemma 5, the identity $\E_{t-1}[\mu_{t-1}^{m_*}(\x_*)] = \E_{t-1}[ f(\x_*)]$ is used without motivation -- which if incorrect would invalidate the proof. Lemma 5 is used to bound term $A_{4,3}$ which is the next issue we will discuss. The function $\mu_{t-1}^{m_*}(\cdot)$ is the posterior mean of a Gaussian process with prior model $m^*$ whereas $f(\x_*) = \sup_{\x \in \X} f(\x)$. For any fixed $m^*$ and $\x$, note that $f(\x) | \x_* = \x$ has a skew-Gaussian distribution \citep{arellano-valleProperties2022}, not a Gaussian distribution, and therefore $f(\x_*)$ does not have the mean $\E[\mu_{t-1}^{m_*}(\x_*)]$ unless the skewness parameter equals 0. We illustrate this with a counterexample at time $t=1$, a known prior (i.e. $M = 1$), and two input locations ($|\X| = 2$). Then, $\E_{t-1}[\mu_{t-1}^{m_*} (\x_*)]$ corresponds to the known prior mean of the optimal input $\x_*$ and $\E_{t-1}[f(\x_*)]$ is the prior mean of $\max_{\x \in \X} f(\x)$.

Consider the bivariate Gaussian distribution $f \sim \mathcal{N} \left( \mathbf{0}, \begin{bmatrix} 1 & \rho \\ \rho & 1\end{bmatrix} \right)$ for $\rho \in (-1, 1)$. Let $f_1$ and $f_2$ denote the first and second entry of $f$. By definition, $\E[f_1] = \E[f_2] = 0$ and therefore $\E_{t-1}[\mu_{t-1}^{m^*}(\x^*)] = \E_{t-1}[0] = 0$.
However, we will show that $\E_{t-1}[f(\x_*)] = \E[\max(f_1, f_2)] > 0$ by direct calculations, contradicting the claimed identity. By the law of total expectation and symmetry of $f_1$ and $f_2$,
\begin{align}
    \E[\max(f_1, f_2)] &= \E[ f_1 | f_1 > f_2] \P(f_1 > f_2) \\
    &\quad+ \E[ f_1 | f_1 \leq f_2] \P(f_1 \leq f_2) \\
    &= \E[f_1 | f_1 > f_2].
\end{align}
By Bayes' rule, symmetry of $f_1$ and $f_2$, and conditional Gaussian update rules, the pdf $p(f_1 = t | f_1 > f_2) = \frac{\P(f_2 < t | f_1 = t) p(f_1 = t)}{\P(f_1 > f_2)}= 2 \phi(t) \Phi\left(t \alpha \right)$ where $\alpha = \sqrt{(1-\rho)/(1+\rho)}$, $\phi(\cdot)$ and $\Phi(\cdot)$ is the pdf and cdf of the unit Gaussian distribution. Using integration by parts, it follows that $\E[\max(f_1,f_2)] = \sqrt{\frac{1-\rho}{ \pi}} > 0$:
\begin{align}
    \E[f_1 | f_1 > f_2] = \int_{\R} 2 t \phi(t) \Phi(t \alpha) dt \\
    = \left[ -2 \phi(t) \Phi(t \alpha) \right]_{-\infty}^{\infty} + \int_\R 2 \alpha \phi(t) \phi(t \alpha) dt \\
    = 0 + \frac{2\alpha}{\sqrt{2\pi}} \int_\R \phi\left(t\sqrt{1+\alpha^2}\right) dt \\
    = \sqrt{\frac{2}{\pi}} \frac{\alpha}{\sqrt{{1+\alpha^2}}} = \sqrt{\frac{1-\rho}{\pi}}.
\end{align}

\subsection{Bound of term $A_{4,3}$ in Eq. (23)} \label{sec:A43}
The term $A_{4,3}= \sum_{t=1}^T \E[2B \I(m_t \notin \C_t)]$ is derived in Eq. (23) by applying Lemma 5 inside the expectation, i.e. it is claimed that $$\E[ (\mu_{t-1}^{m_t}(\x_t) - f(\x_t)) \I (m_t \not\in \C_t)] \leq \E[ 2B \I(m_t \not \in \C_t)]$$ due to Lemma 5. However, applying Lemma 5 is not sufficient to prove this since $\mu_{t-1}^{m_t}(\x_t) - f(\x_t)$ and $\I(m_t \notin \C_t)$ have not been demonstrated to be independent. Given that both terms depend on the selected model $m_t$ (and the history $H_{t-1}$), one would assume the terms to be dependent unless proven otherwise.

\subsection{Step (a) in the bound of term $A_{4,2}$} \label{sec:A42}
In step (a) in the bound of term $A_{4,2} = \sum_{t=1}^T \E[(\mu_{t-1}^{m_t}(\x_t) - f(\x_t)) \I(m_t \in \C_t)]$, it is claimed that $L_{t^m_{\max}}^m(\x) - y_{t^m_{\max}}$ is bounded by $2B$ where $t^m_{\max}$ is defined as ``being the last slot that $m$ is selected". We assume that the authors intended to define $t^m_{\max} = \max \{t \in \{1, \ldots, T\} : m_t = m, m \in \C_t\}$ and claim that 
\begin{equation}
\E[L_{t^m_{\max}}^m(\x_{t^m_{\max}}) - y_{t^m_{\max}}] \leq 2B.
\end{equation}
Assuming Lemma 5 holds, it would be reasonable to assume this claim holds since $L_t^m(\x) \leq \mu_{t-1}^m(\x)$ and $\E[y_t] = \E[f(\x_t)]$ for any fixed timestep $t$. However, since $t_{\max}^m$ is a random timestep that depends on the entire history $H_T$ Lemma 5 cannot directly be applied.

\subsection{Implications for Theorem 2 and 3}
The proof of Theorem 2 states that ``the term $A_4$ can be bounded as in (23)", referring to the step in the proof of Theorem 1 where terms $A_{4,2}$ and $A_{4,3}$ are bounded. In the proof of Theorem 3, the term $C_5$ is the equivalent of $A_4$. Since no motivation for the bound of $C_5$ is provided, we assume it follows from the same reasoning as in Theorem 1. As the proofs of Theorems 2 and 3 rely on the issues discussed for the bounds of $A_{4,2}$ and $A_{4,3}$, the proofs for both theorems are invalid.

\section{Comparison to linear setting of \citet{hongThompson2022}}
Here, we compare the issues identified with the regret bound of EGP-TS with the issues identified by \citet{sandbergAdaptive2026} for MixTS in the linear setting \citep{hongThompson2022}. The proof by \citet{hongThompson2022} introduces an event $E_0$ such that the linear parameter vector is close to its prior mean. In the GP setting, it corresponds to bounding the Mahalanobis distance of $f$ to the true prior distribution $\mathcal{GP}(0, \kappa^{m_*})$. \citet{hongThompson2022} improperly applies the probability-matching property ($x_t | H_{t-1} \eqd x^* | H_{t-1}$ where the history $H_{t-1} = (\x_i,y_i)_{i=1}^{t-1}$) when the event $E_0$ is conditioned on. The proof of \citet{luSurrogate2023} does not introduce the equivalent of $E_0$ and therefore avoids this mistake. 

The event $E_0$ allows \citet{hongThompson2022} to bound the regret by a constant $M$. However, the use of $M$ leads to several incorrect steps, such as bounding the equivalent of $\mu_{t}^{m_t}(x_t) - f(x_t)$ by $M$. \citet{luSurrogate2023} instead bounds the expectation of $\mu_{t}^{m_t}(x_t) - f(x_t)$ in Lemma 5 with an incorrect proof and applies it inside an expectation, as discussed in \cref{sec:lem5,sec:A43}.  

To bound the equivalent of term $A_{4,2}$, \citet{hongThompson2022} surrounds $\mu_{t}^{m_t}(x_t) - f(x_t)$ with $\min(\cdot, M)$ (since $E_0$ has been conditioned on) and subtracts the equivalent confidence radius $\eta \sigma_{t,p_t}(x_t)$ and noise $\epsilon_t$ inside the minimum. This allows for $L_{\tmax}^m(\x) - y_{\tmax}$ to be bounded by $M$. However, adding the noise $\epsilon_t$ inside the minimum decreases the expectation rather than increase it. Therefore, both \citet{luSurrogate2023} and \citet{hongThompson2022} incorrectly bound this term. 

\section{Discussion of revised proof of \citet{luSurrogate2026}}
On May 26 2026, a revision of \citet{luSurrogate2023} was uploaded to arXiv \citep{luSurrogate2026}. The revision addresses the statement of Lemma 3 and the proof of Lemma 5 (\cref{sec:lem3,sec:lem5}) with only a minor modification to the bound in Lemma 5. However, the core issues with the bounds of terms $A_{4,3}$ and $A_{4,2}$ (\cref{sec:A43,sec:A42}) have not been addressed properly. The revised proof contains the similar issues for both terms and consequently the statements of Theorems 1, 2 and 3 have not been properly proven. 
We provide further details on the issues identified in the revised proof below.

\subsection{Revised bound of term $A_{4,3}$} \label{sec:revisedA43}
First, note that the revised Lemma 5 states that $\E[\mu_{t-1}^{m_t}(\x_t) - f(\x_t)] < 3B$, $\forall m_t \in \M, \x_t \in \X, t \in \mathcal{T}$. We assume the inclusion of $\forall m_t \in \M, \x_t \in \X$ in the revised Lemma is a typo since the expectation is unconditional and the proof relies on the distribution of $(m_t, \x_t)$ and not their specific values.

To bound $A_{4,3}$ in Eq (24) in the revised proof \citep{luSurrogate2026}, Lemma 5 is used to show that $\E[ \mu_{t-1}^{m_t}(\x_t) -f(\x_t) | m_t \notin \C_t ] < 3B$. Recall that the original proof \citep{luSurrogate2023} invoked Lemma 5 to bound $\E[ (\mu_{t-1}^{m_t}(\x_t) -f(\x_t)) \I \{ m_t \notin \C_t \} ]$ and note that the revised statement of Lemma 5 differs only in the constant of the bound. 
The objection raised in \cref{sec:A43} is that $\mu_{t-1}^{m_t}(\x_t) -f(\x_t)$ and $m_t \notin \C_t$ are clearly dependent and thus Lemma 5 cannot be applied to $\E[ \mu_{t-1}^{m_t}(\x_t) -f(\x_t) | m_t \notin \C_t ]$ nor $\E[ (\mu_{t-1}^{m_t}(\x_t) -f(\x_t) )\I \{ m_t \notin \C_t \} ]$ without a direct proof that shows that the event $m_t \notin \C_t$ does not affect the distribution of $\mu_{t-1}^{m_t}(\x_t) -f(\x_t)$.

To provide further intuition for why Lemma 5 cannot be applied in this step, note that the revised proof of Lemma 5 relies on the probability matching property of Thompson sampling ($m_t, \x_t | \D_{t-1} \eqd m_*, \x_* | \D_{t-1}$) to show that $\E[\mu_{t-1}^{m_t}(\x_t) | \D_{t-1}] = \E[\mu_{t-1}^{m_*}(\x_*) | \D_{t-1}]$. If we additionally condition on the event $m_t \notin \C_t$, then the distribution of $(m_t, \x_t)$ is altered but the distribution of $(m_*, \x_*)$ is unaffected --- violating the probability matching property.

\subsection{Revised bound of term $A_{4,2}$} \label{sec:revisedA42}
The revised bound of $A_{4,2}$ contains multiple issues that we discuss in turn. We note that the revision clarifies the definition of $\tmax = \max \{t \in \{1, \ldots, T\}: m = m_t \}$. In \cref{sec:A42}, we assumed the authors intended to additionally require that $m \in C_t$.

The second equality in the bound of $A_{4,2}$ splits $\sum_{t=1}^T \E[ (L_{t-1}^{m_t}(\x_t) - y_t) | m_t \in \C_t]$ into a sum over the models $m$: 
\begin{equation}
\begin{aligned}
&\sum_{t=1}^T \E[ (L_{t-1}^{m_t}(\x_t) - y_t) | m_t \in \C_t] \\
&= \sum_{m=1}^M \Big( \E[ L_{\tmax-1}^{m}(\x_\tmax) - y_\tmax | m \in \C_\tmax] 
+ \sum_{t \in \mathcal{T}^m_T \setminus \tmax} \E[ L^m_{t-1}(\x_t) - y_t | m \in \C_t] \Big) \label{eq:sum-over-models}
\end{aligned}
\end{equation}
where $\mathcal{T}^m_T = \{t \in \{1, \ldots, T \}: m_t = m \}$. Since $\mathcal{T}^m_T \setminus \tmax$ is a random set, the right-hand side is stochastic whereas the left-hand side is deterministic.

Step (a) replaces $\sum_{t \in \mathcal{T}^m_T \setminus \tmax} \E[ L^m_{t-1}(\x_t) - y_t | m \in \C_t] $ by $\E[ G^m_\tmax ] = \E[ \sum_{t \in \mathcal{T}^m_T \setminus \tmax}L^m_{t-1}(\x_t) - y_t ]$. Each expectation is conditioned on a different event, so linearity of expectation cannot be used to combine them. In step (b), it is claimed that $\E[G^m_\tmax] \leq \E\left[ 2 \sigma_n \sqrt{(T_m - 1) \log T}\right]$ by definition of $\C_t$. However, since $\tmax$ does not require $m \in C_{\tmax}$, the definition of $\C_t$ cannot be invoked.

For the final-round term, step (a) in Eq (25) claims that $\E[L_{\tmax-1}^m(\x_{\tmax}) - y_{\tmax}  | m \in \C_{\tmax}] \leq 3B$ for any $\tmax$ by Lemma 5 (using that $L_{t}^m(x) \leq \mu_{t-1}^m(\x)$ and $\E[y_t] = \E[f(\x_t)]$). A clarifying sentence states that although $\tmax$ is a random variable, the inequality holds for any value of $\tmax$. There are multiple issues with this claim. 

First, the event $m \in \C_{\tmax}$ is treated as independent of the term in the expectation. Both the lower bound $L_t^m(\x)$ and confidence set $\C_t$ depend on the observations $\D_{t-1}$, thus they cannot simply be treated as independent. Instead, one would expect a detailed proof motivating the use of Lemma 5 under this event. 

Second, the model $m$ is kept fixed but Lemma 5 assumes it is randomly chosen by EGP-TS. As mentioned, whilst Lemma 5 includes the quantifier $\forall m_t \in \M$, the quantifier has no impact on the lemma since the expectation is not conditioned on $m_t = m$. Even if the lemma had included this conditioning, the proof would require revision since it relies on the probability matching property $m_t | \D_{t-1} \eqd m_* | \D_{t-1}$. When $m_t$ is restricted to $m$, this property does not hold. 

Third, the premise of the clarifying sentence, that since the inequality holds for any fixed value $t$ then the expectation must hold for $\tmax$ as well, is false. In mathematical terms, $\E[X_i] \leq B$ $\forall i = 1, ..., N$ does not imply that $\E[X_I] \leq B$ for an arbitrary random index $I \in [N]$. As a counterexample, consider the example provided in \cref{sec:lem5} where $f_1, f_2 \sim \mathcal{N}(0,1)$ and $I = \arg\max_{i \in \{1,2\}} f_i$ with $B = 0$. Then, $\E[f_1] = \E[f_2] \leq 0$ but $\E[f_I] = \E[\max_{i \in \{1,2\}} f_i] > 0$.

\section{Conclusion}
In this paper, we have highlighted issues in the original and revised proofs of \citet{luSurrogate2023,luSurrogate2026} and compared against the issues in the linear setting of \citet{hongThompson2022}. A corrected analysis is given by \citet[Theorem 4.4]{sandbergAdaptive2026}, where EGP-TS is called HP-GP-TS. 

\section*{Acknowledgments}
The work of Jack Sandberg and Morteza Haghir Chehreghani was partially supported by the Wallenberg AI, Autonomous Systems and Software Program (WASP) funded by the Knut and Alice Wallenberg Foundation.

\bibliographystyle{plainnat}
\bibliography{references}

\end{document}